\def\year{2020}\relax
\documentclass[letterpaper]{article} 

\usepackage{amsmath,amsfonts,bm}

\def\eqref#1{equation~\ref{#1}}

\def\1{\bm{1}}

\def\rvx{{\mathbf{x}}}

\def\rvz{{\mathbf{z}}}

\def\mA{{\bm{A}}}

\def\mW{{\bm{W}}}

\DeclareMathAlphabet{\mathsfit}{\encodingdefault}{\sfdefault}{m}{sl}
\SetMathAlphabet{\mathsfit}{bold}{\encodingdefault}{\sfdefault}{bx}{n}

\usepackage{adjustbox}
\usepackage{multirow}
\usepackage{subcaption}
\usepackage{threeparttable}
\usepackage{xcolor}

\usepackage{booktabs}
\usepackage{aaai20}  
\usepackage{times}  
\usepackage{helvet} 
\usepackage{courier}  
\usepackage[hyphens]{url}  
\usepackage{graphicx} 

\usepackage{graphicx}  
\usepackage[symbol]{footmisc}

\title{RTN: Reparameterized Ternary Network}
\author{Yuhang Li\textsuperscript{\rm 1}\thanks{Equal Contribution. }, Xin Dong\textsuperscript{\rm 2}\footnote[1]{}, Sai Qian Zhang\textsuperscript{\rm 2}, Haoli Bai\textsuperscript{\rm 3}, Yuanpeng Chen\textsuperscript{\rm 1}, Wei Wang\textsuperscript{\rm 4}\\   
\textsuperscript{\rm 1}University of Electronic Science and Technology of China \ \ \textsuperscript{\rm 2}Harvard University\\
\textsuperscript{\rm 3}The Chinese University of Hong Kong \ \ \textsuperscript{\rm 4}National University of Singapore\\
\{loafyuhang, haolibai,chenyuanpengcyp\}@gmail.com, \{xindong, zhangs\}@g.harvard.edu, wangwei@comp.nus.edu.sg 
}

\begin{document}
\maketitle
\begin{abstract}
To deploy deep neural networks on resource-limited devices, 
quantization has been widely explored.
In this work, we study the extremely low-bit networks which have tremendous speed-up, memory saving with quantized activation and weights.
We first bring up three omitted issues in extremely low-bit networks:
the squashing range of quantized values; the gradient vanishing during backpropagation and the unexploited hardware acceleration of ternary networks. 
By reparameterizing quantized activation and weights vector with full precision scale and offset for fixed ternary vector, 
we decouple the range and magnitude from the direction 
to extenuate the three issues. 
Learnable scale and offset can automatically adjust the range of quantized values and sparsity without gradient vanishing. 
A novel encoding and computation pattern are designed to support efficient computing for our reparameterized ternary network~(RTN). 
Experiments on ResNet-18 for ImageNet demonstrate that the proposed RTN finds a much better efficiency between bitwidth and accuracy, and achieves up to 26.76\% relative accuracy improvement compared with state-of-the-art methods. Moreover, we validate the proposed computation pattern on Field Programmable Gate Arrays~(FPGA), and it brings $46.46\times$ and $89.17\times$ savings on power and area respectively compared with the full precision convolution. 
\end{abstract}

\maketitle

\section{Introduction}

Deep neural networks have achieved significant improvement for various real-world applications.
However, the large memory cost, computational burden, and energy consumption prohibit the massive deployment of deep neural networks on resource-limited devices.
A number of methods are proposed to compress and accelerate deep neural networks, including pruning~\cite{han2015deep}, tensor decomposition~\cite{zhang2015accelerating}, and quantization~\cite{rastegari2016xnor}.

Among these methods, low-bit network quantization is particularly helpful in network acceleration and size reduction.
Binary neural networks~\cite{courbariaux2015binaryconnect} raise a lot of attention. 
However, binary networks usually suffer from a large drop in terms of accuracy due to limited expressiveness. 
To enhance the model capacity, various multi-bit quantization methods are proposed~\cite{zhou2016dorefa}, 
which significantly improve the performance of quantized models but enjoy less size reduction and speed acceleration.

As a compromise between binary networks and \textit{N}-bit networks, 
ternary neural networks convert full-precision parameters into merely three values and save a large amount of memory with acceptable accuracy degradation. 
Despite ternary networks are popularly investigated~\cite{li2016twn,zhu2016ttq} in recent years, three major issues are mostly overlooked: 
1) \textit{The squashing behavior of the forward quantization function.}
Most existing activation quantization methods~\cite{cai2017hwgq,rastegari2016xnor} squash full precision activation values into a narrow and fixed range, which could affect the expressiveness of the quantized network.
2) \textit{The saturating behavior of the backward quantization function.}
The clipped Straight-Through Estimator~(STE)~\cite{bengio2013estimating} is widely adopted in training a quantized network. Nevertheless, the gradient becomes zero when entering the saturating zone of the STE estimator. Moreover, as the network depth increases, the training could suffer from the severe problem of gradient vanishing. 
3) \textit{Hardware customization for ternary neural networks.}
For networks with ternary weights and activation values, the computation on most modern hardware can only be performed when the ternary values are 2-bit aligned. 
Compared with 2-bit quantization, 
it is yet less explored to utilize some nice properties of ternary values to design a more efficient computation pattern and save more energy.

In this paper, we propose a reparameterized ternary network (RTN) to resolve the three issues. Specifically, in RTNs both weights and activation values are ternarized, followed by a reparameterization with scale and offset parameters.
The reparameterization can easily alleviate the first two issues mentioned above.
Specifically, in order to avoid the squashing behavior of quantization function during the forward pass, the learnable scale and offset parameters 
on network parameters 
enable dynamic adjustment of the quantization range and thereon enhances the capacity of the ternary network.
To tackle the saturating behavior of the clipped STE function, with the chain rule of derivatives we can decompose the gradient of activation after reparameterization with respect to that of activation before reparameterization, as well as the gradients of scale and offset parameters. 
Consequently, even though the gradient of an activation before reparameterization saturates, the optimization can still proceed 
as a result of learning the reparameterization parameters.

Finally, to address the third issue, we build a customized hardware prototype on FPGA for the reparameterized ternary network. We design an efficient encoding and computation pattern to conduct dot products between two ternary vectors, saving extra energy compared to the existing implementations of 2-bit networks.

Experimental results on large scale tasks like ImageNet indicate that our proposed method significantly improves the capacity of the ternary network, and achieves up to $26.76\%$ relative improvement of accuracy on ResNet-18 against state-of-the-art binary and low-bit networks. Moreover, our hardware prototype on FPGA achieves $3.43\times$ and $4.17\times$ savings on power and area respectively comparing to traditional implementations of the 2-bit network.

\section{Related Work}
Recent work on network compression shows that full precision computation is not necessary for the training and inference of DNNs~\cite{gupta2015deep}. 
To achieve higher compression and acceleration ratio, extremely low-bit like binary weights~\cite{rastegari2016xnor} have been studied. 
\cite{li2016twn,zhu2016ttq} further improve the performance by ternarizing weights to achieve higher representation ability. TWN minimizes the Euclidean distance between ternary weights and the full precision weights. Instead of the symmetric ternarization, TTQ uses an asymmetric ternarization to achieve higher performance but less hardware convenience. 

Substantial speed up requires further quantization for activation, which is generally more challenging than weights quantization~\cite{cai2017hwgq}. \cite{courbariaux2016bnn} uses $+1$ and $-1$ to represent both weights and activation and XNOR-Net~\cite{rastegari2016xnor} further adds scaling factors for binary weights to improve accuracy. 
Higher-order Residual Quantization~\cite{li2017performance} uses two 1-bit tensors to approximate the full precision activation, but the computation speed would reduce to half. To take advantage of ReLU~\cite{nair2010relu} and introduce sparsity in quantized activation, \cite{cai2017hwgq} uses Half-wave Gaussian Quantization to approximate ReLU. The quantized activation function has the form of a step-wise function, which always has zero gradients with respect to its input. To circumvent this problem, Straight-Through Estimator~(STE)~\cite{bengio2013estimating} is adopted. STE approximates backward function of arbitrary functions with (clipped)~identity function, and several studies~\cite{liu2018birealnet,zhou2016dorefa} attempt to reduce this mismatch between forward and backward to improve performance. \cite{choi2018pact,baskin2018nice} propose to learn the clipping parameters and achieve better results. \cite{gong2019dsq} leverage $tanh$ function to approximate the gradient of quantization, however, there is still a large accuracy gap between extremely low-bit and full precision models.

\section{Methodology}
\label{sec:methodology}

For a weight filter in a convolution layer, it is denoted by $\mW\in \mathbb{R}^{c\cdot k\cdot k}$, 
where $c$ and $k$ are the number of input channels and the kernel size, respectively. Suppose one instance is fed to the network, and the corresponding feature map is denoted by $\mA\in\mathbb{R}^{c\cdot k\cdot k}$.
Then the output of one unit 
in the next layer can be computed by the dot product\footnote{For convolutional layers, this can be done by the im2col operation.} as 
    $z=\phi(\mW^T \mA)$,
where $\phi(\cdot)$ is the Rectified Linear Units~\cite{nair2010relu}.

\begin{figure}
  \centering
  \includegraphics[width=\linewidth]{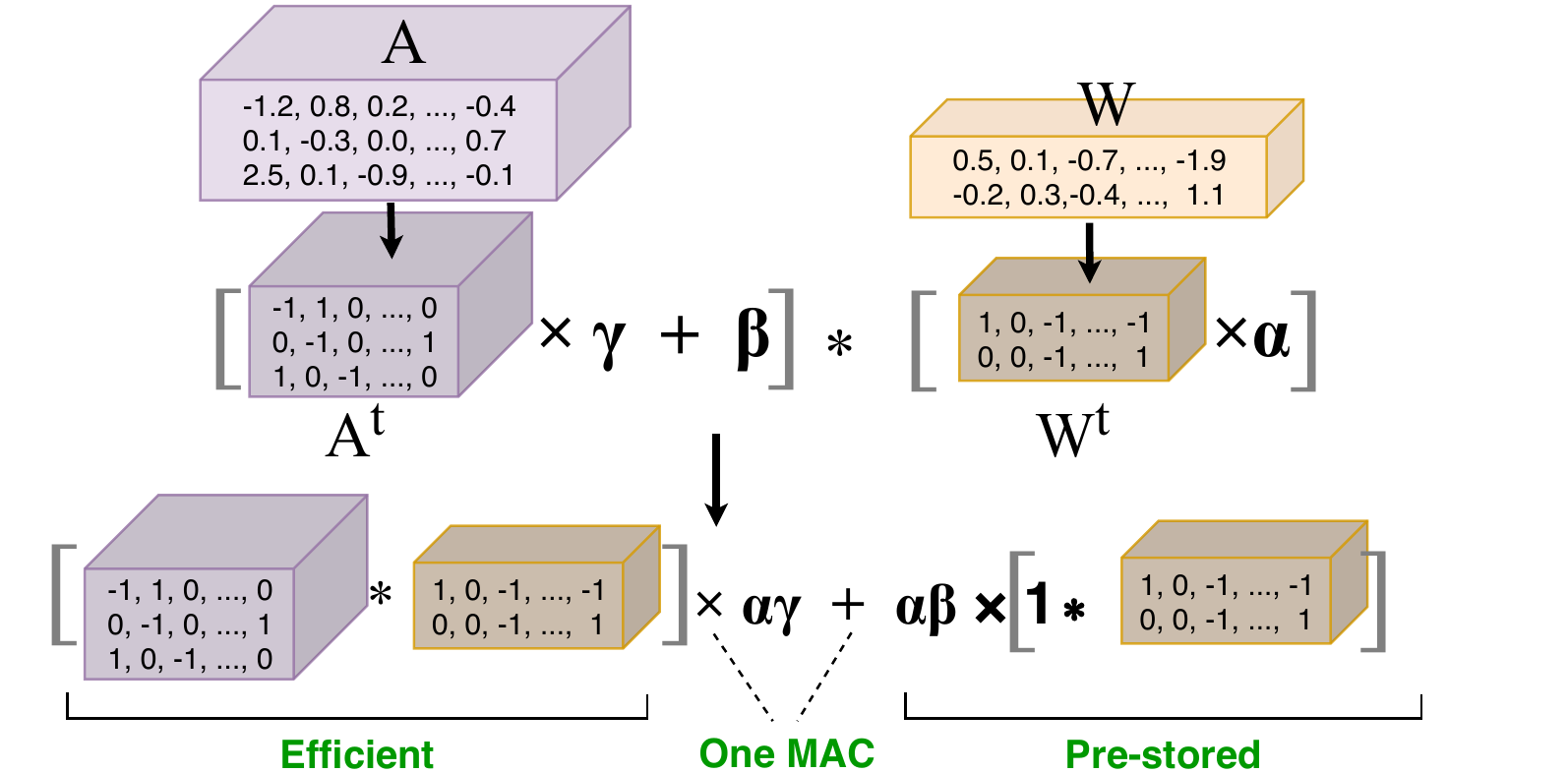}
  \caption{The overall workflow of our proposed reparameterized ternary quantization. The weights and activation are reparameterized by a scale (and an offset) factor to achieve higher network capacity. The convolution is efficient with our specially designed computation pattern.}
  \label{fig:intro-scale-offset}
\end{figure}

Our proposed reparameterized ternary network (RTN) consists of the linear transformations on both weights and activation values of the network. The reparameterization allows the dynamic adjustment of the quantization range, and avoid the issue of gradient vanishing during the quantized training. 
Besides, we also customize hardware implementations for RTN by leveraging the nice properties of ternary networks.
The overall workflow of RTN is shown in Figure~\ref{fig:intro-scale-offset}.

\subsection{Reparameterized Ternarization}
\label{sec:act_ternarization}

\subsubsection{Activation Ternarization}

Previous work~\cite{wan2018tbn,deng2018gxnor} on ternary networks argue that the degradation of quantized network mainly comes from the limited quantization levels.
However, it is rarely observed that the quantization functions they adopt usually squash the input into fixed ranges and therefore harm the network expressiveness significantly. 
In a ternary neural network, the quantization function is applied to both weights and activations, which highly restricts the capacity of the quantized model.
Therefore, in this paper, we propose a reparameterized quantizer to enhance the model expressiveness.
First, the ternarization function is given by:
\begin{equation}
    \mA_i^t= Q(\mA_i)=
    \begin{cases}
    \text{sign}(\mA_i) &  \text{if    }|\mA_i|>0.5\\
    \ 0 &  \text{otherwise }
    \end{cases},
\label{eqn:activation_quant}
\end{equation}
Since activation function in RTN is ReLU, the output of this function is always non-negative, which means activations can never be quantized to $-1$. We apply Batch Normalization (BN) after ReLU to recreate negative activations. As a result, the quantization values can be made full use of.

After normalizing the inputs of each layer, BN usually applies an affine transformation to increase the model capacity. Here we use 
\begin{equation}
\label{equ:activation-transformer}
    \bar \mA = k\mA + b
\end{equation} to denote the transformation. With BN transformation, consequently the quantization function can be formulated as follows:
\begin{equation}
    \mA_i^t= Q(\bar \mA_i)=
    \begin{cases}
    1 &  \text{if }\mA_i>\frac{0.5-b}{k} \\
    -1 &  \text{if }\mA_i<-\frac{0.5+b}{k}\\
    0 & \text{otherwise }
    \end{cases},
\label{eqn:activation-quant-transform}
\end{equation}
where the learnable BN parameters $k$ and $b$ can adaptively adjust the quantization threshold~($0.5$) in Equation~\ref{eqn:activation_quant}. In spite of the quantization threshold is learnable, however, the ternary activation $\mA_i^t$ only contains fixed ternary values~(i.e. $\mA^t\in\{+1, 0, -1\}^n$).
We consider a ternary activation $\bar\mA^t\in\{-\gamma+\beta,\beta, \gamma+\beta\}$
and we further reparameterize $\bar\mA^t$ by
\begin{equation}
    \bar{\mA}^t=\gamma\cdot\mA^t+\beta,
\label{equ:act-reparameterize}
\end{equation}
where $\gamma$ is the magnitude scale factor and $\beta$ is the offset. With $\gamma$ and $\beta$, the reparameterized ternary activation can dynamically change the squashing range, improving the network capacity with little increase of model size and computation. Here, we refer to $\mA^t$ as fixed ternary activation, because the ternary values are fixed and it only controls the direction of the activation vector.

Our method can be reduced to a number of previous methods by taking different $\beta$ and $\gamma$.
For example, when $\beta = \gamma$, we squash the activation into range $[0, \beta+\gamma]$, which is equivalent to HWGQ~\cite{cai2017hwgq}.
When $\gamma=\mathbb{E}_{|\mA|>0.5}(|\mA|)$ and $\beta=0$, the Euclidean distance from $\bar\mA^t$ to full precision activation $\mA$ is minimized, and our approach resembles XNOR-Net~\cite{rastegari2016xnor}.
Note that the scale factor mentioned in XNOR-Net is different from ours, their scale factor has to be calculated from full precision activations for each forward pass as a running variable, which is not practical.\footnote{As a result they abandon this scale factor for quantized activation in the officially released implementation.}
A similar idea on decoupling the vector magnitude from its direction
for full precision weights can also be found in~\cite{salimans2016weightnormalization}. Note that these factors are designed in a layerwise pattern, so value ranges may change across different layers depending on $\gamma$ and $\beta$.

\subsubsection{Weight Ternarization}
In a similar spirit to activation ternarization, we first apply linear transformation for network weights that resembles the BN layer in activation to obtain learnable quantization thresholds.
For each weight filter $\mW\in\mathbb{R}^{c\cdot k\cdot k}$, the weights transformer is defined as follows:
\begin{equation}
\label{equ:weight-transform}
    \bar \mW = k_{\mW}\mW+b_{\mW},
\end{equation}
where $k_{\mW}$ and $b_{\mW}$ are learnable parameters. Then the transformed weights $\bar \mW$ are quantized by the same function in Equation~\ref{eqn:activation_quant}, i.e $\mW^t=Q(\bar\mW)$. As a consequence, the weights can adjust its quantization threshold.
To obtain flexible quantized values, we follow a similar way to reparameterize $\mW^t$ by $\bar\mW^t = \alpha \mW^t \in\{-\alpha, 0, \alpha\}$, where $\alpha$ is the scale factor. Note that 
the offset is not included under the consideration of additional computation overhead.

\subsection{Backward Update in Reparameterized Ternarization}
\begin{figure}
\centering
  \includegraphics[width=0.75\linewidth]{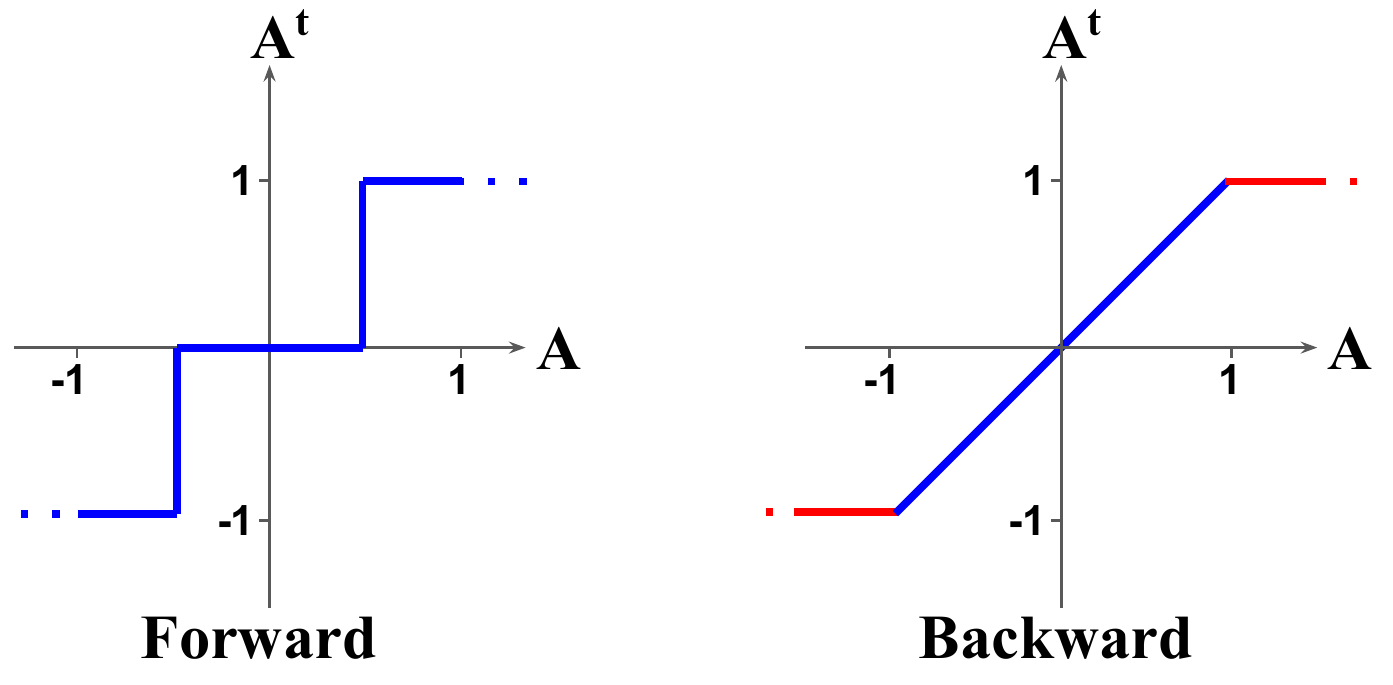}
  \caption{The forward and backward functions for fixed ternary activation $\mA^t = Q(\mA)$. The line in red is referred to as saturating zone whose gradient is always zero.}
  \label{fig:saturated}
 
\end{figure}

A typical approach to propagate the gradients through the quantized activation is the clipped Straight-Through Estimator (STE): $\frac{\partial \mA^t}{\partial \bar\mA}=\mathbf{1}_{|\bar\mA|\leq 1}$, 
which is exactly the gradients of \emph{hard tanh}. Despite being successfully used in previous methods~\cite{rastegari2016xnor}, \emph{hard tanh} suffers from the saturating problem. When $|\bar\mA_i|\ge 1$, the gradient of $\bar\mA_i$ becomes zero, which enters the saturating zone as shown in the red part of Figure~\ref{fig:saturated} . The saturating behavior of STE can cause gradient vanishing for weights as the depth of the network increases, which slows down and even hurts the convergence of the model. Furthermore, once activation falls into the saturating zone, they will get stuck and barely find a way out because both $\mW, \mW^t$ and $\mA,\mA^t$ remain unchanged. 

Fortunately, our reparameterized ternary activation can alleviate this problem easily. Consider $L$ as the loss function, the derivative w.r.t. to $\bar\mA^t$ can be written as
\begin{equation}
\label{equ_grad_Y}
\frac{\partial L}{\partial \gamma}= \mA^t \frac{\partial L}{\partial\bar\mA^t},\  \frac{\partial L}{\partial \beta}= \frac{\partial L}{\partial\bar\mA^t}.
\end{equation}
It can be observed that since we decouple the scale $\gamma$ and offset $\beta$ from fixed ternary activation $\mA^t$, even when $|\bar\mA|\ge 1$ the reparameterized ternary activation $\bar\mA^t$ can still be optimized as a result of learning $\gamma$ and $\beta$. 
Consequently, the entire network can converge faster and reach a better optimum in the loss landscape. 

Furthermore, our reparameterized ternarization has another benefit that it can dynamically adjust the learning rate of network parameters. 
Consider the gradients w.r.t to the activation,
\begin{equation}
\label{eqn:lr_adjust}
    \frac{\partial L}{\partial \bar\mA}=\frac{\partial L}{\partial \bar\mA^t}\frac{\partial \bar\mA^t}{\partial \bar\mA}=\gamma\mathbf{1}_{|\mA|\leq 1}\frac{\partial L}{\partial \bar\mA^t},
\end{equation}
where $\gamma$ can be absorbed in learning rate as a multiplier, making the training of the network robust to the value of the learning rate. 
Learnable scale factor also has been studied in ~\cite{salimans2016weightnormalization,zhu2016ttq}
, in which they claim a similar effect as well.

\subsection{The XOR-XNOR Toy Problem}
\label{sec:toy-problem}
\begin{figure}
\centering
  \includegraphics[width=0.95\linewidth]{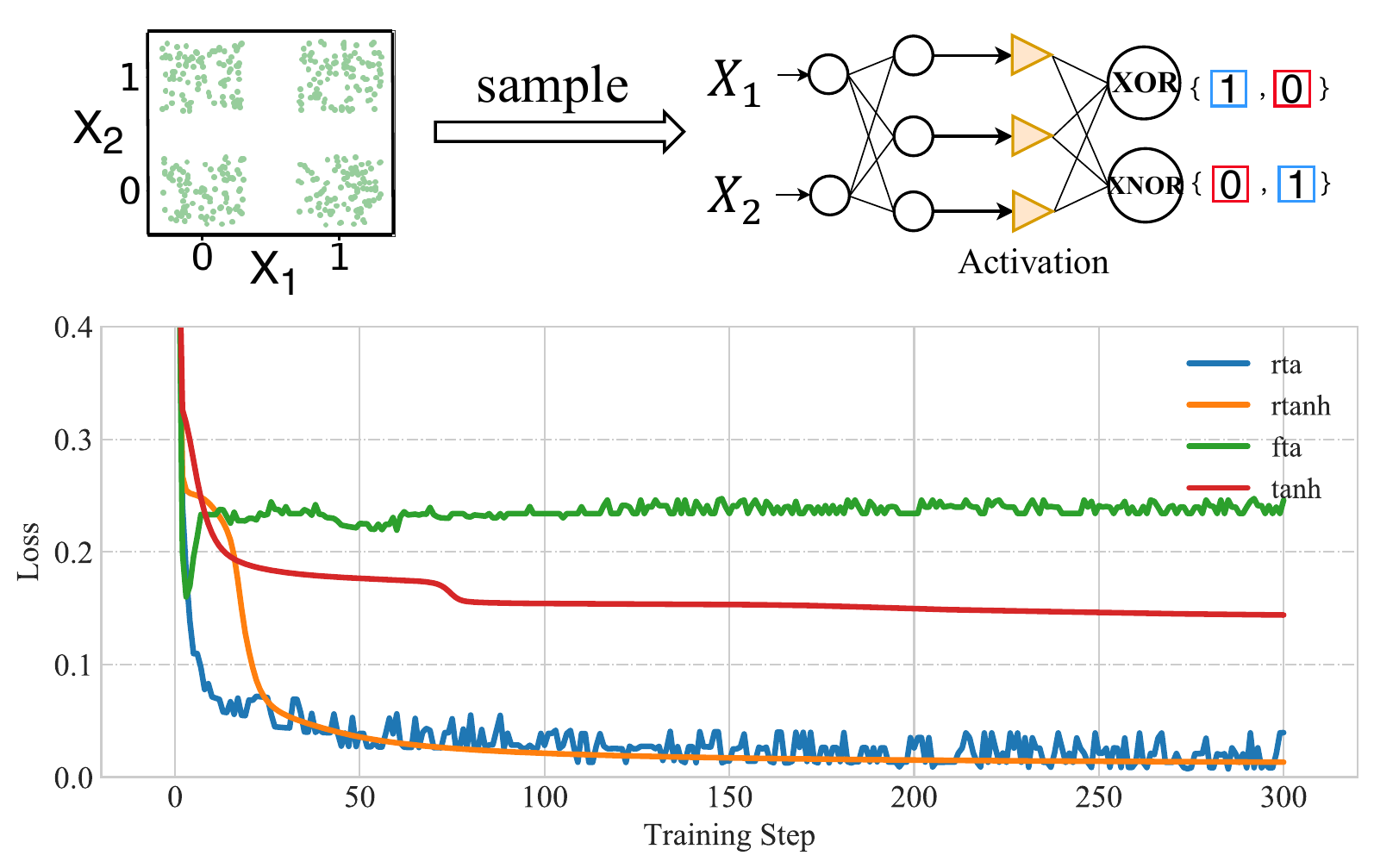}
  \caption{The XOR-XNOR toy model architecture and the training curve for the toy model.}
  \label{fig:toy_loss}
\end{figure}

To demonstrate how the reparameterized ternarization improves the capacity of the quantized model, we give a toy example on a two-layer neural network, as is shown in Figure~\ref{fig:toy_loss}. 

The 2-layer network is designed to learn two logical functions, $\text{XOR}(\rvx_1,\rvx_2)$ and $\text{XNOR}(\rvx_1,\rvx_2)$ respectively. 4 different kinds of activation function are compared: fixed ternary activation ($fta$), 
reparameterized ternary activation ($rta$), the hyperbolic tangent activation ($tanh$) and the reparameterized hyperbolic tangent activation~($rtanh$).
Inputs are sampled from a Bernoulli distribution plus a uniform noise, $\{(\rvx_1=\rvz_1+\mathbf{\epsilon}_1, \rvx_2=\rvz_2+\mathbf{\epsilon}_2)|\rvz_1, \rvz_2\sim\mathcal{B}(p=0.5), \mathbf{\epsilon}\sim \mathcal{U}(-0.3, 0.3)\}$. 
Outputs are either 0 or 1.
The network has a hidden layer consisting of 3 neurons without the bias term. To better observe behaviors of quantized activation, we keep the weights as full precision numbers.
We report the mean square error~(MSE) during training.
More implementation details are in the Appendix. 

The training curve is shown in Figure~\ref{fig:toy_loss}. 
Compared with fixed ternary activation (\emph{fta}) and reparameterized ternary activation (\emph{rta}), hyperbolic tangent~(\emph{tanh}) is a full precision function with a fixed squashing range $[-1,+1]$, which is  supposed  to have better representation capability than the ternary activation functions.
However, our \emph{rta} achieves lower MSE than \emph{tanh}, because the scale and offset factors alleviate the squashing issue. Similarly, \emph{rtanh} achieves lower MSE than \emph{tanh} and \emph{rta}.
In particular, the scale and offset factors of \emph{rtanh} are $\gamma^*=0.46$ and $\beta^*=2.02$ respectively, which substantially change the squashing range from $[-1, 1]$ into $[1.56, 2.48]$. From the empirical result we can see the range of activation values is at least as important as the number of quantization levels.


\begin{table} 
\centering
\caption{This table shows the bit encoding scheme of our ternary values and 2-bit quaternary values.}
\footnotesize
\begin{tabular}{cccc}
\hline
\multicolumn{2}{c}{2-bit Representation}&  &   \\ \cline{1-2}
1st bit & 2nd bit & \multirow{-2}{*}{\begin{tabular}[c]{@{}c@{}}Our Ternary\\ True Value\end{tabular}} & \multirow{-2}{*}{\begin{tabular}[c]{@{}c@{}}2-bit Network\\ True Value\end{tabular}} \\ \hline
0    &0 & 0  & 0  \\
0    & 1  & 0  & 1  \\
1   & 0   & -1 & 2 \\
1    & 1  & +1  & 3 \\ \hline
\end{tabular}
\label{table:bit-encoding}
\end{table}
\subsection{Efficient Computation Pattern}
\label{computation pattern}

\subsubsection{How to Compute Dot Product between Two Ternary Vectors}
To support our ternary network~(ternary weights + ternary activations), in this section, we propose an efficient way to compute the dot product between the ternary weights and activation vectors, which is the core operation for both convolution and linear layers. A special bit encoding scheme is adopted for the ternary weights and activations. We use two bits to represent each ternary weight and activation, where the first bit indicates whether this number is zero or not, and the second bit indicates the sign of this number. Table~\ref{table:bit-encoding} shows the detailed encoding scheme for all ternary values $-1$, $0$ and $+1$. 
Under this encoding scheme, zero can be represented by either $00$ or $01$.

\label{sec:computation pattern}

Now, we have $\mW^t,\mA^t\in\{+1,0,-1\}^{ck^2}$ and we will encode them into 2-bit vector representations. Suggest $\mW_1\in\{0,1\}^{ck^2}$ is a vector contains the first bit of all entries in ternary weights. $\mW_2$ contains the second bit and we define $\mA_1, \mA_2$ in a similar way to represent the activation. The dot product can be computed using bit-wise operations.
\begin{equation}
    ({\mW^t})^T{\mA^t}=bC\left ({\bold{c}}\right)-2\times bC\left(({\mW_2} \oplus {\mA_2})\wedge {\bold{c}} \right),
    \label{eqn:convert-conv}
\end{equation}
where ${\bold{c}}={\mW_1}\wedge{\mA_1}$, and $\wedge \text{ and }\oplus$ are AND, XOR bit-wise operations respectively. $bC(.)$ returns the number of 1 (logic high) in a vector. As indicated by Equation~\ref{eqn:convert-conv}, the convolution can be computed efficiently via simple Boolean operations.

\begin{figure}
\centering
  \includegraphics[width=0.95\linewidth]{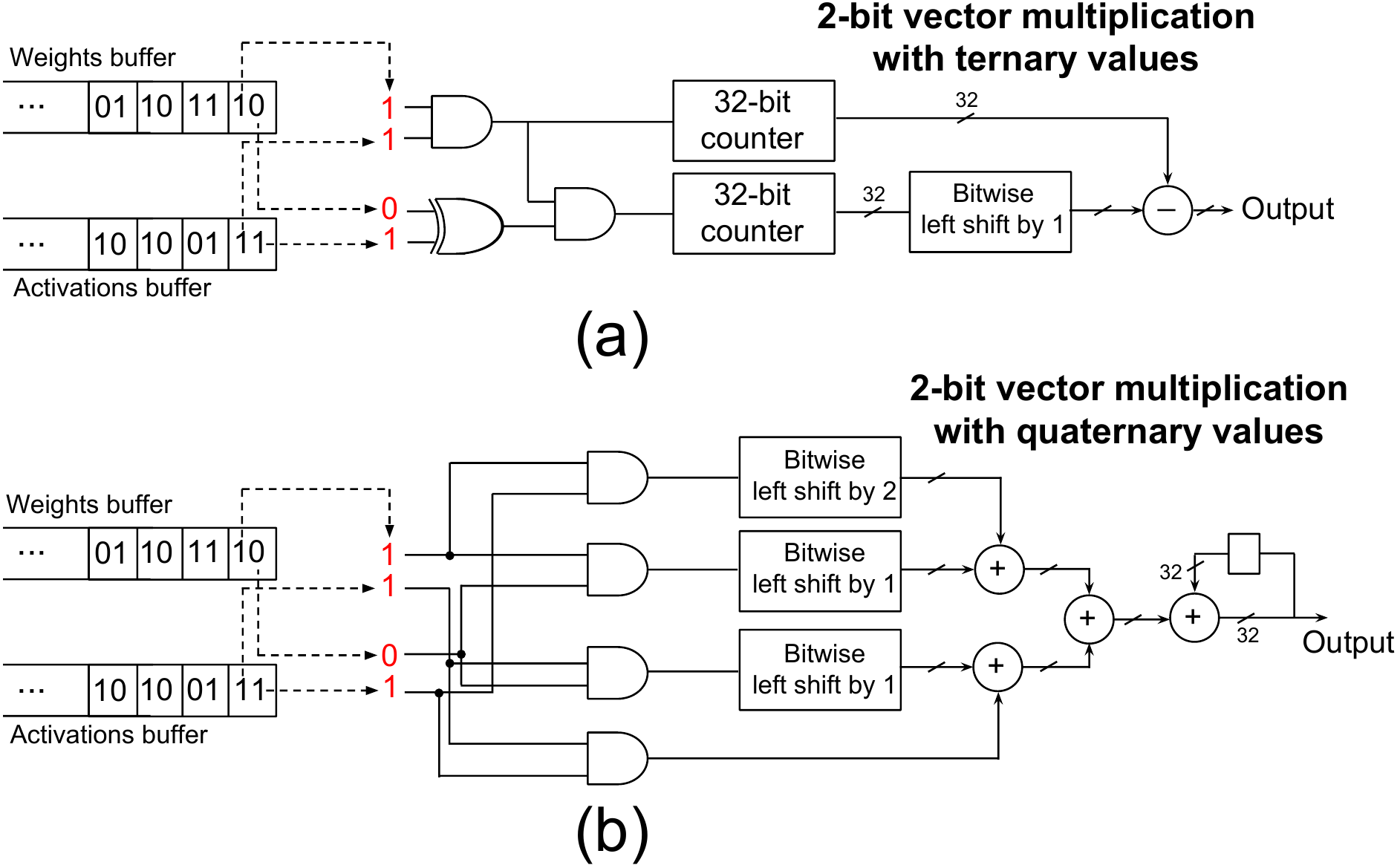}
  \caption{This figure illustrates the hardware implementation of the dot product between, (a) ternary weights and ternary activations (b) 2-bit weights and 2-bit activations.}
  \label{fig:conv-circuit}
 
\end{figure}

Figure~\ref{fig:conv-circuit}(a) shows the hardware design for the vector multiplication shown in Equation~\ref{eqn:convert-conv}. Given the two input vectors, the circuit computes the scale product between each pair of elements of the two vectors, the partial results $bC\left ({\bold{c}}\right)$ and $bC\left(({\mW_2} \oplus {\mA_2})\wedge {\bold{c}} \right)$ are saved inside two 32-bit counters. The multiplication with two shown in Equation~\ref{eqn:convert-conv} can be easily achieved by shifting the partial results to the left by 1 bit. A substractor is used to perform the substraction operation shown in Equation~\ref{eqn:convert-conv}.

For comparison, we compute the dot product of 2-bit quaternary weights and activation because they share the same size of our model. 
We use the computation pattern introduced in DoReFa-Net~\cite{zhou2016dorefa} to compute the dot product.

The quaternary vector multiplication can be computed by performing AND between each bit of inputs.
Figure~\ref{fig:conv-circuit}(b) shows the hardware design for the vector multiplication in~\cite{zhou2016dorefa}. The circuit takes the two input vectors and calculates the scale product between each pair of elements. The multiplication with power-of-two is implemented by using bitwise shift operations. Four adders are employed to sum the partial results. We also evaluate the performances of the two designs shown in Figure~\ref{fig:conv-circuit} in terms of power consumption, computation latency, and area in the next section.  

\subsubsection{How To Deal With $\gamma$ and $\beta$}
Actually, our reparameterized ternary activation has two extra parameters, scale $\gamma$ and offset $\beta$, besides fixed ternary activation. We demonstrate that it only introduces negligible computation complexity. With the quantized ternary weights $\alpha\mW^t$ and reparameterized ternary activation $\bar\mA^t$, the input of the next layer can be computed by
\begin{equation}
\begin{aligned}
    z &=\phi\left (\alpha\mW^t*\bar\mA^t\right)=\phi\left (\alpha\mW^t*(\gamma\mA^t+\beta)\right)\\
             &=\phi\left (\alpha\gamma(\mW^t\otimes\mA^t)\ +\ C\right),\ \ \ 
    C = \alpha\beta(\bold{1}\otimes\mW^t),
\end{aligned}
\label{equ:combine-compute}
\end{equation}
where $\phi$ is the ReLU function, $\otimes$ is the dot product between ternary vectors, $\bold{1}$ denotes the matrix with all elements equal to 1. As a matter of fact, the second term, i.e., $C$, in Equation~\ref{equ:combine-compute}, is a constant, which can be pre-stored in the cache. As shown in Figure~\ref{fig:intro-scale-offset}, when performing the convolution, we first calculate the ternary value convolution efficiently with  Boolean operations, then we only need to conduct one multiply-accumulate (MAC) operations to get the final results. 
\subsubsection{Reparameterized Ternary Activation Can Adjust Sparsity Automatically}
Interestingly, we can modify the expression of Equation~\ref{equ:combine-compute} and fold the second term into ReLU to make it more hardware friendly,
\begin{equation}
    z=\phi\left (\alpha\gamma(\mW^t\otimes\mA^t)\ +\ C\right) = \alpha\cdot\phi_{T}\left (\gamma(\mW^t\otimes\mA^t)\right),
\label{eqn:rta with relu}
\end{equation}
where $\phi_{T}(\boldsymbol{x})=\max(0, \boldsymbol{x}+T)$
is the ReLU parameterized by the sparsity threshold $T = \beta(\bold{1}\otimes\mW^t)$. 

Apparently, $T$ controls the sparsity of $z$. This reveals another effect of our reparameterized ternary activation. It can control the sparsity of the activation. The sparsity of activation has been studied in~\cite{wang2018twostep}, in which they find that sparsity has a profound impact on accuracy. However, \cite{wang2018twostep} manually sets the sparsity threshold to increase the sparsity, in which they believe the quantization error can be reduced and larger activation is more important based on the attention mechanism. In our method, the sparsity threshold is given by $\beta(\bold{1}\otimes\mW^t)$, which can be dynamically tuned during the training by offset factor $\beta$ for every layer. We give the sparsity record in Section ~\ref{results-imagenet} to show that our method concurs with~\cite{wang2018twostep}.

\begin{figure*}[t]
    \centering
    \includegraphics[width=\textwidth]{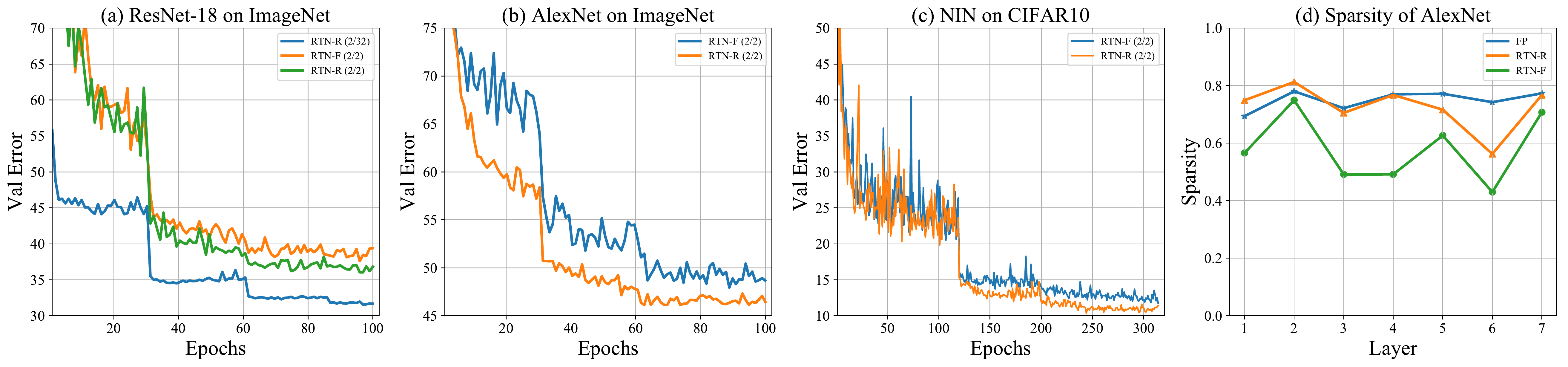}
    \caption{The validation error rate plots during training with/without reparameterization and the sparsity comparison of each layer in AlexnNet.}
    \label{fig:plots}
\end{figure*}
\section{Experiments}
\label{sec:experiment}

In this section, we first present some empirical evaluations of the reparameterized ternary network (RTN) on two real-world datasets: ImageNet-ILSVRC2012~\cite{russakovsky2015imagenet} and CIFAR-10~\cite{krizhevsky2009learning}, then we evaluate the performance of the hardware implementation for RTN in terms of power consumption and area.

We adopt a number of popular neural architectures for evaluation: ResNet~\cite{he2016resnet}, AlexNet~\cite{alexnet}, MobileNet~\cite{howard2017mobilenets} and Network-In-Network (NIN)~\cite{lin2013nin}. 
Two sets of strong baselines are chosen for comparison:
1) quantizing weights only:
BWN~\cite{rastegari2016xnor}, TWN~\cite{li2016twn}, and TTQ~\cite{zhu2016ttq};
2) quantizing both weights and activations: 
XNOR~\cite{rastegari2016xnor}, Bi-Real~\cite{liu2018birealnet}, TBN~\cite{wan2018tbn}, HWGQ~\cite{cai2017hwgq}, DoReFa-Net~\cite{zhou2016dorefa}, PACT~\cite{choi2018pact} and HORQ~\cite{li2017horq}.

We denote our method with (resp. without) reparameterization on weights and activations as RTN-R (resp. RTN-F). We also evaluate our method when only weights are quantized.

We highlight substantial accuracy improvement~(\textbf{up to 13\% absolute improvement} compared with XNOR-Net) of our RTN for ResNet-18 on ImageNet. Such improvement mainly comes from: 1) zero is introduced into quantized activation to get the fixed ternary activation~$\{-1, 0, +1\}$, 2) dynamically adjusting the quantization range of weights and activations by Equations (\ref{equ:activation-transformer}) and (\ref{equ:weight-transform}), and 
3) learnable scale and offset are adopted for the fixed ternary activation to get the reparameterized ternary activations~$\{\gamma+\beta, \beta, -\gamma+\beta\}$ which have much better representation capability with negligible computation overhead.

Compared with several 2-bit models, RTN has the lowest degradation from full precision models and achieves comparable accuracy. In Section~\ref{sec:hardware-imple}, we implement our ternary multiplication circuit and other 2-bit multiplication circuit used in~\cite{zhou2016dorefa,choi2018pact,li2017horq} and show that the circuit for multiplication with ternary values significantly outperforms that for multiplication with 2-bit values in terms of power and area. 


\subsection{Implementation}

We follow the implementation setting of other extremely low-bit quantization networks~\cite{rastegari2016xnor} and do not quantize the weights and activation in the first and the last layers. 
See Appendix for more details of our implementation.

Initialization could be vitally important for quantization neural networks. We first train a full-precision model from scratch and initialize the RTN 
by minimizing the Euclidean distance between quantized and full precision weights like TWN~\cite{li2016twn}. For example, the initial $\gamma$ is set to $\mathbb{E}_{|\mA|>0.5}(|\mA|)$ and $\beta$ is set to 0.

\subsection{Results on ImageNet}
\label{results-imagenet}

The validation error on ImageNet is plotted in Figure~\ref{fig:plots}. We can see that RTN-R has a lower error rate than RTN-F. 
Especially, AlexNet plot, Figure~\ref{fig:plots}(b), shows that RTN-R has a relatively smooth curve and better convergence speed and may be the result of automatic adjustment of learning rate via gamma in Equation~\ref{eqn:lr_adjust}.

The overall results on ImageNet are shown in Table~\ref{table:overall-result} with several strong extremely low-bit models.
Note that we swap the order of BN and ReLU, so we report the full precision models' accuracy as a reference and compare the degradation from full precision models (the last column in the table).
We first compare our RTN with models that only quantize weights like BWN, TWN, and TTQ. 
Our RTN not only achieves state-of-the-art accuracy but also has the smallest gap between full precision models. 
In addition, compared with TTQ's asymmetric quantization, our RTN uses symmetric quantization which is naturally harder for training but more friendly for hardware implementation. 

\begin{table}[]
\caption{
Overall comparison of various extremely low-bit quantized models on ImageNet and CIFAR10. We compare top-1 accuracy and degradation from full precision for a fair comparison. $^\dag$ Denotes network uses ternary values instead of quaternary for 2 bits representation; $^\ddag$ denotes results of ResNet-18B where the filter number in each block is 1.5$\times$.}
\centering
\small
\begin{tabular}{l c c c c}
\toprule
\multicolumn{5}{c}{\textit{ResNet-18 (ImageNet)}}\\
\midrule
Methods & \# bits(W/A) & FP ref. & Accuracy & Degrad. \\
\hline
BWN & 1 / 32 & 69.3 & 60.8 & 8.5 \\
$\text{TWN}^{\dag}$ & 2 / 32 & 69.3 & 61.8 & 7.5 \\
$\text{TWN}^{\dag\ddag}$ & 2 / 32 & 69.3 & 65.3 & 4.0 \\
$\text{TTQ}^{\dag\ddag}$ & 2 / 32 & 69.6 & 66.6 & 3.0 \\
$\text{RTN-R}^{\dag}$ & 2 / 32 & 69.2 & \textbf{68.5} & \textbf{0.7} \\
\hline
XNOR & 1 / 1 & 69.3 & 51.2 & 18.1\\
Bi-Real & 1 / 1 & 68.0 & 56.4 & 11.6 \\
$\text{TBN}^{\dag}$ & 1 / 2 & 69.3 & 55.6 & 13.7 \\
DoReFa & 1 / 2 & 70.2 & 53.4 & 16.8 \\
HWGQ & 1 / 2 & 69.6 & 56.1 & 13.5 \\
HORQ & 2 / 2 & 69.3 & 55.9 & 13.4 \\
DoReFa & 2 / 2 & 70.2 & 62.6 & 7.6 \\
PACT & 2 / 2 & 70.2 & 64.4 & 5.8 \\
$\text{RTN-F}^{\dag}$ & 2 / 2 & 69.2 & 62.4 & 6.8\\
$\text{RTN-R}^{\dag}$ & 2 / 2 & 69.2 & \textbf{64.5} & \textbf{4.7}\\
\midrule
\multicolumn{5}{c}{\textit{AlexNet (ImageNet)}}\\
\midrule
XNOR & 1 / 1 & 56.6 & 44.2 & 12.4\\
$\text{TBN}^{\dag}$ & 1 / 2 & 57.2 & 49.7 & 7.5 \\
DoReFa & 1 / 2 & 55.9 & 49.8 & 6.1 \\
HWGQ & 1 / 2 & 55.7 & 50.5 & 5.2 \\
PACT & 2 / 2 & 57.2 & \textbf{55.0} & \textbf{2.2} \\
$\text{RTN-F}^{\dag}$ & 2 / 2 & 58.7 & 52.6 & 6.1 \\
$\text{RTN-R}^{\dag}$ & 2 / 2 & 58.7 & 53.9 & 4.8 \\
\midrule
\multicolumn{5}{c}{\textit{MobileNet (ImageNet)}}\\
\midrule
PACT & 2 / 2 & 69.9 & 56.1 & 13.8 \\
$\text{RTN-R}^{\dag}$ & 2 / 2 & 69.9 & \textbf{56.9} & \textbf{13.0} \\
\midrule 
\multicolumn{5}{c}{\textit{NIN (CIFAR10)}}\\
\midrule
XNOR & 1 / 1 & 89.8 & 86.4 & 3.4\\
$\text{RTN-F}^{\dag}$ & 2 / 2 & 89.8 & 88.2 & 1.6\\
$\text{RTN-S}^{\dag}$ & 2 / 2 & 89.8 & 88.5 & 1.3\\
$\text{RTN-O}^{\dag}$ & 2 / 2 & 89.8 & 89.1 & 0.7\\
$\text{RTN-R}^{\dag}$ & 2 / 2 & 89.8 & \textbf{89.6} & \textbf{0.2}\\
\bottomrule
\end{tabular}
\label{table:overall-result}
\end{table}

Quantizing the activation is more challenging compared with weights~\cite{cai2017hwgq}, and there is still a large margin between full precision models and extremely low-bit models. 
We compare several models that quantize both weights and activation
with our proposed model~(denoted by RTN-R).
For the ablation study, we also report the performance of our ternary network with fixed ternary activation~(denoted by RTN-F) to show the effectiveness of scale and offset.

According to Table~\ref{table:overall-result}, we can conclude that, \textbf{1)}  RTN-R outperforms almost every models. So, though it is a tradeoff between the number of bits and accuracy, the ternary network finds a better balance between them.
\textbf{2)} In spite of PACT has comparable performance, especially on AlexNet, our RTN-R also shares a small gap with full precision models. Note that RTN is, furthermore, better for hardware implementation on mobile and embedded devices.
\textbf{3)} With learnable scale factors and offset, RTN-R has higher accuracy than the RTN-F, which validates the improvement of representation ability from our reparameterization design.

\subsubsection{Sparsity Comparison}
According to our analysis in Section ~\ref{sec:computation pattern}, the offset $\beta$ can adjust the sparsity of $z$ automatically. Generally, changing of sparsity concurs with observation in~\cite{wang2018twostep}, in which they believe the optimal sparsity is slightly higher than 50\% based on the foundation of attention mechanism. Figure~\ref{fig:plots}~(d) shows the sparsity comparison between RTN-F, RTN-R, and full precision models.
Our reparameterized ternary activation can adjust the sparsity automatically, and the sparsity of RTN-R is close to FP. Compared with~\cite{wang2018twostep}, 
Our RTN can adjust sparsity automatically without any manual settings.

\subsubsection{Analysis of Reparameterization}
We report the value of scale $\gamma$ and offset $\beta$ for activation and the mean value of scale $\bar\alpha$ for weights of each layer in ResNet-18 (See Appendix). We can see that the activation distribution has changed a lot among layers. This means that each layer learns its optimal range and magnitude thus increasing the representational ability. Interestingly, we found that the activation and weights in the downsample residual layer only change slightly. This situation may result from the special $1\times1$ filter in this layer.

\begin{figure}[t]
    \centering
    \includegraphics[width=1.05\linewidth]{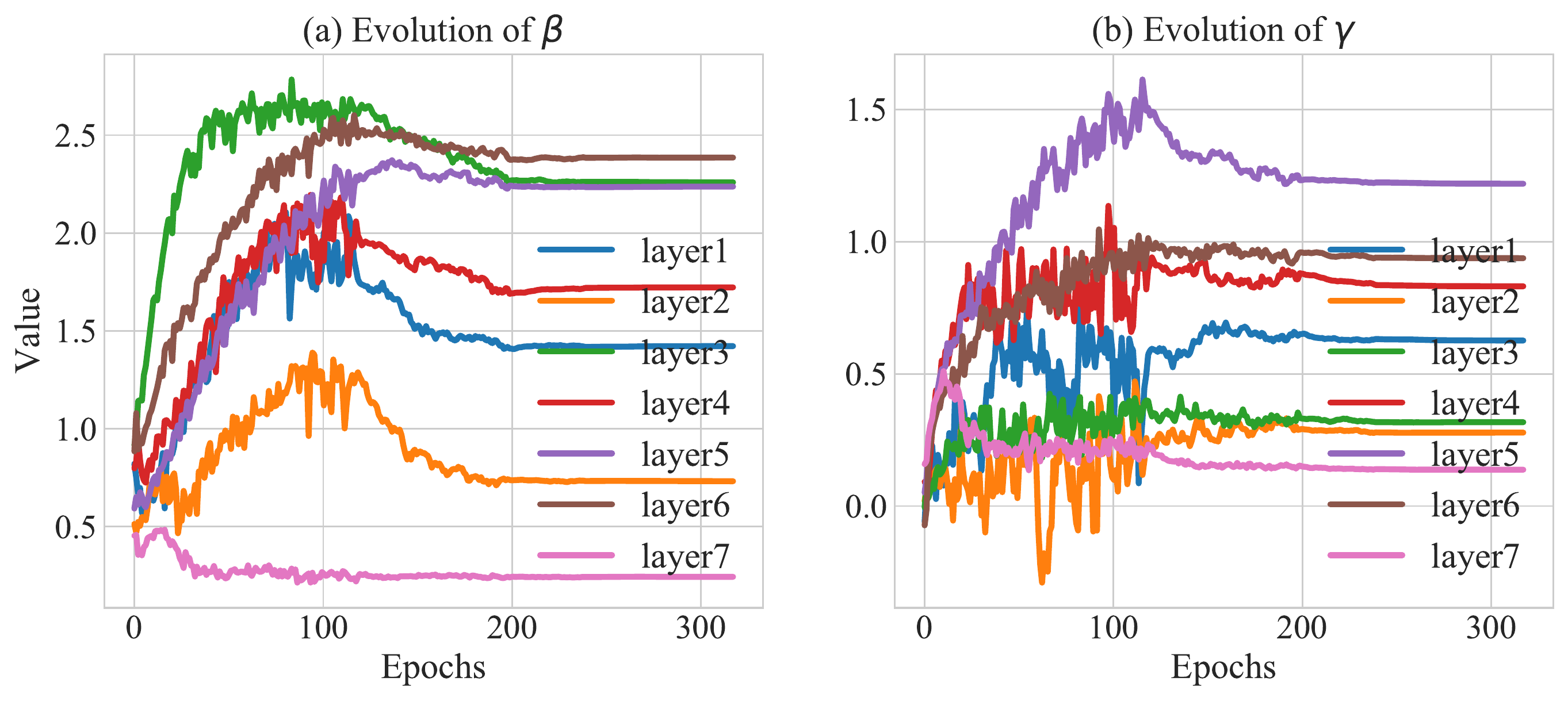}
    \caption{The evolution of $\gamma$ and $\beta$ in NIN on CIFAR10.}
    \label{fig:gamma}
\end{figure}
\label{sec:results-CIFAR}

\subsection{Results on CIFAR10}
For CIFAR10, we mainly compare our method with XNOR-Net on NIN.
We use the PyTorch implementations of XNOR-Net~\cite{rastegari2016xnor}\footnote{\url{https://github.com/jiecaoyu/XNOR-Net-PyTorch}}. See more implementations details in Appendix.

Results for NIN on CIFAR10 can be found in Table~\ref{table:overall-result}. Our RTN almost reboots full precision accuracy~(only 0.2\% absolute gap) without bells and whistles. This performance may result from the scale and offset that significantly changes the range of ternary activation and weights. 

\subsubsection*{Evolution of $\gamma$ and $\beta$}
In Figure~\ref{fig:gamma}, we show the evolution of the parameters in RTA. For $\gamma$, almost all of them will increase at the beginning of the training, and are downscaled when the learning rate is decreased. For $\beta$, they are more volatile. Nonetheless, the evolution of these parameters are directly optimized by training objective and the heuristic is not easy to predict.


\subsubsection*{Ablation Study}
There are two learnable parameters in the reparameterized ternary activation, the scale factor, and the offset factor. We evaluate the effect of these two parameters by only applying one of them in the RTN. We denote the RTN-S as the activation with the scale factor and RTN-O as the activation with the offset only. Implementation on CIFAR10 is kept the same as before.

The results are shown in Table~\ref{table:overall-result}. Apparently, when we only add the scale factor, the improvement can be trivial. 
This is because the ReLU does not impact the scale of the activation (i.e. $\phi(\gamma\mA)=\gamma\phi(\mA)$), and BN can eliminate the effect of the scale factor. We refer to this effect as \textit{scale invariance of activation}. However, according to Equation~\ref{eqn:rta with relu}, the offset factor can change the sparsity threshold in ReLU, thus greatly affect the activation. Therefore, RTN-O has higher performance than RTN-S. Note that in our RTN-R, there is no \textit{scale invariance of activation} when we apply scale and offset factors together, which can both change the distribution of activation.

\subsection{Hardware Implementation}
\label{sec:hardware-imple}
We compare the hardware performances of the two circuits for vector multiplication operation shown in Figure~\ref{fig:conv-circuit}. We further implement the circuit for floating-point values (32 bits) vector multiplication. We synthesize our design with Xilinx Vivado Design Suite~\cite{vivado} and use Xilinx VC707 FPGA evaluation board for power measurement. For the comparison on circuit area and computation latency, we utilize the Synopsys Design Compiler~\cite{synopsysdesigncompiler} with 45nm NanGate Open Cell Library~\cite{nangatelib}. 

\begin{table}
\caption{Hardware performances of the circuits for the vector multiplication operations shown in Figure~\ref{fig:conv-circuit}.}

 \label{table:hardware-comparison}
\scriptsize
\centering
 \begin{adjustbox}{width=\columnwidth,center}
 \begin{tabular}{|c |c |c|}
 \hline
  Circuits & Power consumption & Area\\
 \hline
  2-bit vector  &     &     \\
  multiplication with  & \textbf{22.17$uW$} & \textbf{199.43$um^2$}   \\
  ternary values  &     &  \\
  \hline
   2-bit vector  &     &     \\
  multiplication with  & 76.09$uW$ & 831.62$um^2$  \\
  quaternary values  &     &  \\
   \hline
  vector multiplication &     &     \\
   with floating-point & 1.03$mW$ & 17783$um^2$  \\
   (32 bits) values   &     &  \\
 \hline
 \end{tabular}
 \end{adjustbox}

\end{table}

As shown in the Table~\ref{table:hardware-comparison}, the circuit for ternary values (Figure~\ref{fig:conv-circuit}(a)) outperforms that for the 2-bit values~(quaternary values) (Figure~\ref{fig:conv-circuit}(b)) and floating-point values in terms of both power (3.43$\times$, 46.46$\times$) and area $(4.17\times,89.17\times)$. These differences result from the fact that more adders and bitwise shifters are used by the circuit for quaternary value multiplication. From Table~\ref{table:hardware-comparison}, we notice that the circuit for quaternary value multiplication is $4\times$ larger than that of ternary value multiplication. That is to say, for a fixed size of circuit area and a settled clock frequency, the circuit for ternary value multiplication has $4\times$ less latency than the circuit for quaternary value multiplication, since we can make four ternary value multiplier works in parallel. Moreover, our circuit can be easily deployed as a building block of any large-scale parallel computing framework such as systolic array~\cite{kung1982systolic} for efficient matrix multiplication.
\section{Conclusion}

In this paper, we propose the reparameterized ternary network with ternary weights and activation. The learnable reparameterizers are demonstrated to considerably increase the expressiveness of fixed ternary values. According to our analysis and empirical results, scale and offset are able to adjust the range of quantized value, inflect sparsity of activation and accelerate training. To support efficient computing in RTN, a novel computation pattern is proposed. 
\paragraph{Acknowledgements}
This work is supported by the National Research Foundation, Prime Ministers Office, Singapore under its National Cybersecurity
RD Programme (No. NRF2016NCR-NCR002-020), and FY2017 SUG Grant.

\bibliographystyle{aaai} 
\bibliography{sample-base}

\appendix

\section{Experimental Details}
\subsection{Layer Order}
\begin{figure}[h]
\centering
 \includegraphics[width=0.83\linewidth]{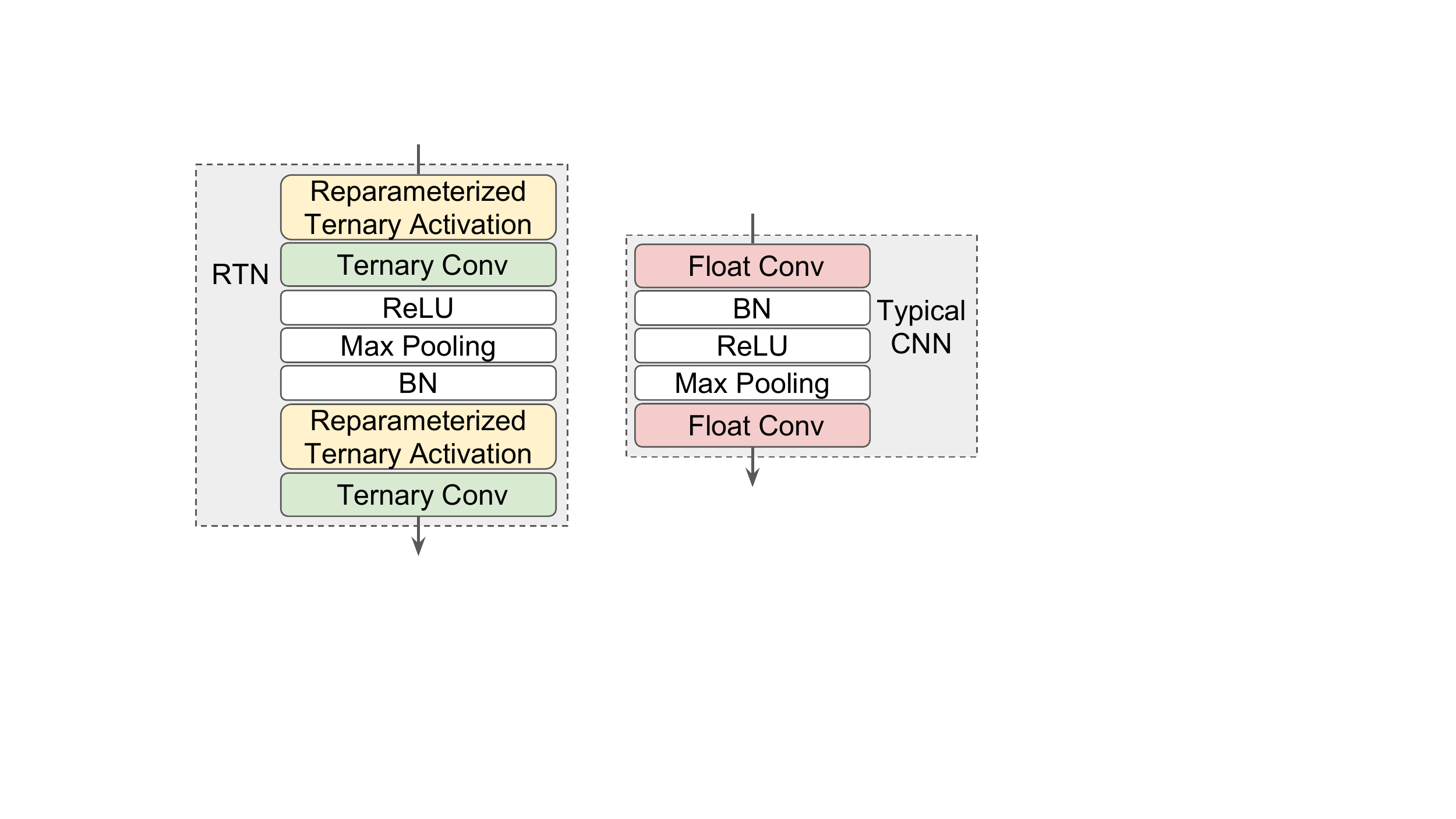}
 \caption{This figure illustrates the basic block structure in our proposed RTN (right) and the typical CNN (left). Our model swap the order of ReLU and BN to better ternarize activation.} 
 \label{fig:layer-order}
\end{figure}

In RTN, we swap the order of BN and ReLU, the details are shown in Figure~\ref{fig:layer-order}.

\subsection{Implementation Details of XOR-XNOR toy problem}
The 2-layer network is designed to learn two logical functions, $\text{XOR}(\rvx_1,\rvx_2)$ and $\text{XNOR}(\rvx_1,\rvx_2)$ respectively.
Inputs are sampled from a Bernoulli distribution pulsing a uniform noise, $\{(\rvx_1=\rvz_1+\mathbf{\epsilon}_1, \rvx_2=\rvz_2+\mathbf{\epsilon}_2)|\rvz_1, \rvz_2\sim\mathcal{B}(p=0.5), \mathbf{\epsilon}\sim \mathcal{U}(-0.3, 0.3)\}$. 
Outputs are either 0 or 1.
The network has a hidden layer consisting of 3 neurons without bias term. To better observe behaviors of quantized activation, we keep the weights as full precision numbers.

We use 4 different kinds of activation function for comparison, which are fixed ternary activation ($fta$), reparameterized ternary activation ($rta$), the hyperbolic tangent activation ($tanh$) and the reparameterized hyperbolic tangent activation(i.e. $tanh$ with scale and offset factor)($rtanh$). 
Except for the $rtanh \text{ and } rta$, the other two activation are squashing non-linearity in a fixed range $(-1,1)$. 
Both $rta\text{ and }fta$ have limited~(only 3) quantization levels. 
The hyperbolic tangent is full-precision, which is supposed to have better representation ability than ternary activation. 
We use MSE loss and stochastic gradient descent to train the network. 
The learning rate is 0.03 and we train the toy model for 15000 epochs.



\subsection{Implementation Details of Main Experiments}
For the ImageNet dataset, training images are randomly resized and cropped randomly to $224 \times 224$.
randomly 256$\times$256 on the smaller dimension and then a random crop of 224$\times$224 is selected for training. 
Training images are horizontally flipped in a random way. The test images are centrally cropped to $224\times 224$ (227 for training and test images in AlexNet).
We use Stochastic Gradient Decent (SGD) as optimizer. Weight decay is set as 0.0001 for ResNet-18 and AlexNet. 
Each network was trained up to 100 epochs with batch size of 1024. Learning rate starts from 0.1 and is decayed by a factor of 10 at epoch 30,60,85. For quantization parameters (e.g. $\gamma,\beta$ for activation and $k_{\mW}, b_{\mW}, \alpha$ for weights), we should set a lower learning rate because their gradients are a summation over each elements in weights/activation which increases its magnitude. In practice, we find that 0.001 is appropriate for weight quantization parameters and 0.1 for activation quantization parameters. 

As for NIN on CIFAR10, we use Adam as parameter optimizer and we train the network for 320 epochs. The weight decay was set to 0.00001, and the initial learning rate is 0.01 with a decrease factor of 0.1. Note that we also do not quantize the first and the last layer.

\begin{table}[]
\caption{Scale~($\gamma,\bar\alpha$) and offset~($\beta$) of ResNet-18 on ImageNet. Note that we use $\bar\alpha$ to denote the mean $\alpha$ in all weights filters across one layer.}
\centering
\begin{tabular}{c r r r}
\hline
Layer & \multicolumn{1}{c}{$\gamma$} & \multicolumn{1}{c}{$\beta$} & \multicolumn{1}{c}{$\bar{\alpha}$} \\
\hline
layer1.0.conv1 & 1.0426 & $-$0.0308 & 1.8160 \\
layer1.0.conv2 & 0.9729 & $-$0.2344 & 1.0974 \\
layer1.1.conv1 & 1.0223 & 0.1699 & 2.0325 \\
layer1.1.conv2 & 0.7962 & 0.0956 & 1.6872 \\
layer2.0.conv1 & 1.3083 & 0.5152 & 3.0458 \\
layer2.0.conv2 & 0.8191 & 0.6840 & 1.5639 \\
layer2.0.downsample.0 & 1.0000 & $-$0.0024 & 0.8739 \\
layer2.1.conv1 & 1.4091 & 0.3080 & 1.4284 \\
layer2.1.conv2 & 0.7678 & 0.4921 & 2.3644 \\
layer3.0.conv1 & 1.3986 & 0.8014 & 2.7552 \\
layer3.0.conv2 & 0.8916 & 0.7033& 1.6015 \\
layer3.0.downsample.0 & 0.9996 & 0.0000& 0.9435 \\
layer3.1.conv1 & 1.6719 & 0.4738 & 2.9345 \\
layer3.1.conv2 & 1.0112 & 0.4731 & 2.1110 \\
layer4.0.conv1 & 2.0472 & 1.4202& 3.0216 \\
layer4.0.conv2 & 1.1033 & 0.9717 & 1.7116 \\
layer4.0.downsample.0 & 1.0037 & 0.0000 & 0.8537 \\
layer4.1.conv1 & 2.4687 & 1.4774& 1.8244 \\
layer4.1.conv2 & 0.8959 & 0.7186& 2.3379\\
\hline
\end{tabular}
\label{tab:scale and offset}
\end{table}

\end{document}